\documentclass{article}
\usepackage{spconf,amsmath,graphicx}
\usepackage{amssymb}
\usepackage{booktabs}
\usepackage{subcaption}
\usepackage[hidelinks]{hyperref}
\usepackage{cleveref}
\usepackage{xcolor}

\title{DYNAMIC DISTILLATION AND GRADIENT CONSISTENCY FOR ROBUST LONG-TAILED INCREMENTAL LEARNING}
\name{Taigo Sakai$^{\star}$ \qquad Kazuhiro Hotta$^{\star}$}
\address{$^{\star}$ Meijo University, 1-501 Shiogamaguchi, Tempaku-ku, Nagoya 468-8502, Japan}

\begin{document}
\ninept
\maketitle
\begin{abstract}
The task of Long-tailed Class Incremental Learning (LT-CIL) addresses the sequential learning of new classes from datasets with imbalanced class distributions. This scenario intensifies the fundamental problem of catastrophic forgetting, inherent to continual learning, with the dual challenges of under-learning minority classes and overfitting majority classes.
To tackle these combined issues, this paper proposes two main techniques. First, we introduce gradient consistency regularization, which leverages the moving average of gradients to suppress abrupt fluctuations and stabilize the training process. Second, we dynamically adjust the weight of the distillation loss by measuring the degree of class imbalance with normalized entropy. This adaptive weighting establishes an optimal balance between retaining old knowledge and acquiring new information.
Experiments on the CIFAR-100-LT, ImageNetSubset-LT, and Food101-LT benchmarks show that our method achieves consistent accuracy improvements of up to 5.0\%. Furthermore, we demonstrate dramatic gains in the challenging 'In-ordered' setting, where tasks progress from majority to minority classes, highlighting our method's robustness in mitigating forgetting under unfavorable learning dynamics. This enhanced performance is achieved without a significant increase in computational overhead, demonstrating the practicality of our framework.
\end{abstract}
\section{INTRODUCTION}
\label{sec:intro}

Deep learning has demonstrated remarkable advancements in visual recognition and language understanding~\cite{lecun2015deep}. However, real-world applications require incremental model updates as new data becomes available. Consequently, Class Incremental Learning (CIL) has emerged as an active research focus~\cite{10599804, ijcai2024p924}. The primary challenge in CIL is catastrophic forgetting, where the integration of new tasks overwrites previously acquired representations~\cite{Goodfellow2013AnEI, forget, MCCLOSKEY1989109}. This problem is exacerbated by uneven task ordering and data distributions, presenting challenges not encountered in traditional batch learning.

Furthermore, real-world data often exhibits a long-tailed distribution, containing a limited number of majority classes and numerous minority classes. Incremental learning under these conditions, known as Long-Tailed Class Incremental Learning (LT-CIL), has recently gained increased attention~\cite{liu2022long, zhang2023deep}. LT-CIL combines catastrophic forgetting with the dual challenges of under-learning minority classes and overfitting majority classes.

Existing LT-CIL methods often employ complex strategies, such as sub-prototype allocation or two-stage training. While effective, these approaches demand large models, extended data retention, and lengthy retraining, raising privacy and efficiency concerns. Therefore, developing memory- and computationally-efficient methods is required.

The proposed methodology builds upon Gradient Reweighting (GR), which dynamically adjusts gradient magnitudes to balance class recognition. However, basic GR exhibits substantial gradient fluctuations that delay convergence. To overcome this, gradient consistency regularization is introduced to stabilize learning by constraining abrupt gradient changes via a moving average. This prevents excessive parameter shifts, strengthening minority class representations while stabilizing the learning of new classes. Furthermore, knowledge distillation~\cite{Hinton2015DistillingTK} is dynamically adjusted using normalized entropy to counteract class imbalance. Because majority class predictions can dominate distillation loss, normalized entropy adaptively scales distillation strength. This prevents majority classes from overshadowing new class learning, enabling flexible, distribution-aware knowledge retention.

Experimental results demonstrated up to 5.0\% accuracy improvements over standard CIL (iCaRL, PODNet) and imbalanced CIL (GR) methods~\cite{rebuffi2017icarl, Douillard2020PODNetPO, He_2024_CVPR}. {Although dedicated LT-CIL methods exist, GR was selected as a baseline because it achieves highly competitive accuracy. The primary objective is overcoming the performance ceiling in the 'In-ordered' scenario, where classes are introduced sequentially from majority to minority. Although baseline methods yield numerically higher scores in the In-ordered setting compared to the Shuffled setting, their performance is bounded by severe overfitting to early majority classes, which restricts the recognition of subsequent minority classes.} The proposed framework bypasses this limitation by attenuating initial majority-class bias and stabilizing gradients, strictly improving intra-scenario performance.
Particularly, GCR notably stabilizes training, enhancing feature extraction for minority classes. The proposed methodology maintains high accuracy without extensive data retention, making it suitable for environments with limited memory and computation.

The remainder of this paper is structured as follows. Section 2 discusses related works on continual learning and class imbalance. Section 3 details our proposed method, including adaptive distillation coefficients and gradient consistency regularization. Section 4 presents experimental setups and results on CIFAR100-LT, ImageNetSubset-LT, and Food101-LT. Finally, Section 5 concludes the paper and discusses future research directions.

\section{RELATED WORKS}
\label{sec:Related}

\subsection{Class-Incremental Learning}

Continual Learning is a setting where models sequentially learn multiple tasks that become available over time, rather than having access to all data at once. Class-Incremental Learning (CIL), a particular form of continual learning, focuses on maintaining classification accuracy across all classes as new classes are incrementally introduced. Thus, in CIL, the model must learn new classes without prior knowledge of future classes, while preserving recognition performance for previously learned ones. Under these constraints, the phenomenon of catastrophic forgetting, in which the integration of new classes leads to overwriting of previously learned knowledge, frequently occurs. Consequently, the central challenge in CIL is balancing the flexibility to incorporate new classes with the stability needed to retain existing knowledge.
One straightforward solution to catastrophic forgetting is replay-based methods, which preserve a portion of previous task data for reuse in subsequent learning stages. Prominent examples, such as iCaRL~\cite{rebuffi2017icarl}, ER~\cite{NEURIPS2019_fa7cdfad}, LMCL~\cite{10350694}, and Effective Generative Replay~\cite{YANG2025113477}, utilize stored samples to facilitate the relearning of past distributions. While effective, these approaches present practical limitations, requiring extensive storage and computation time. Moreover, storing raw data indefinitely may be infeasible due to privacy or legal restrictions in fields like healthcare. Thus, replay-based methods necessitate careful consideration of storage, computational load, and privacy constraints.

To overcome these limitations, several methods have been developed to minimize stored data volume while retaining previous task knowledge. Specifically, these approaches store feature vectors or statistical summaries, such as mean and variance per class, compressing historical information to facilitate memory-efficient learning and privacy preservation~\cite{mai2021supervised}. Knowledge distillation is another established approach for alleviating forgetting. This methodology uses the outputs, specifically soft targets, from past tasks as supervisory signals, encouraging models to replicate previous responses when learning new tasks. Methods such as iCaRL~\cite{rebuffi2017icarl}, {PODNet~\cite{Douillard2020PODNetPO},} LwF~\cite{li2017learning}, and others concurrently optimize classification loss and distillation loss, thus maintaining accuracy on old classes while improving recognition of new ones~\cite{Hou_2018_ECCV}. Currently, knowledge retention through distillation has been adopted as a standard protocol in many CIL methods.

% \subsection{Class-Incremental Learning in Long-tailed scenario}

However, real-world datasets often exhibit class imbalance, particularly with long-tailed distributions where sample sizes significantly vary among classes. This imbalance further complicates continual learning because models naturally overfit to abundant majority classes, severely reducing accuracy on minority classes. Techniques to address long-tailed distributions, such as cost-sensitive learning, where losses are weighted per class, and oversampling methods, including synthetic data generation (e.g., SMOTE, ADASYN), have long been known~\cite{menon2021longtaillearninglogitadjustment,chawla2002smote,he2008adasyn, Yu-Hang_2025_CVPR}. Nevertheless, directly applying these methods to CIL often leads to interference between old and new classes, exacerbating forgetting. Consequently, Long-Tailed Class Incremental Learning (LT-CIL) requires new integrated frameworks beyond those designed solely for either CIL or traditional long-tailed learning.

\subsection{Gradient Reweighting}
\label{GR_DAKD}

Gradient Reweighting (GR){~\cite{He_2024_CVPR}} dynamically adjusts gradient weights to enhance minority class learning without storing past data. However, naive GR is susceptible to instability due to gradient fluctuations. While GR often employs Distribution-Aware Knowledge Distillation to mitigate forgetting, balancing new and old knowledge remains challenging~\cite{qwae083}. Alternative approaches, including sub-prototype spaces~\cite{10654831} or two-stage training~\cite{liu2022long,sraghavan2024delta,qi2025adaptive}, address these issues but often increase memory and computational costs. To overcome these limitations, the proposed methodology integrates GR with Gradient Consistency Regularization (GCR) and dynamic distillation, ensuring stable and efficient learning under imbalanced conditions.

\section{PROPOSED METHOD}
\label{sec:Method}

\begin{figure*}[t]
    \centering
    \includegraphics[width=0.9\linewidth]{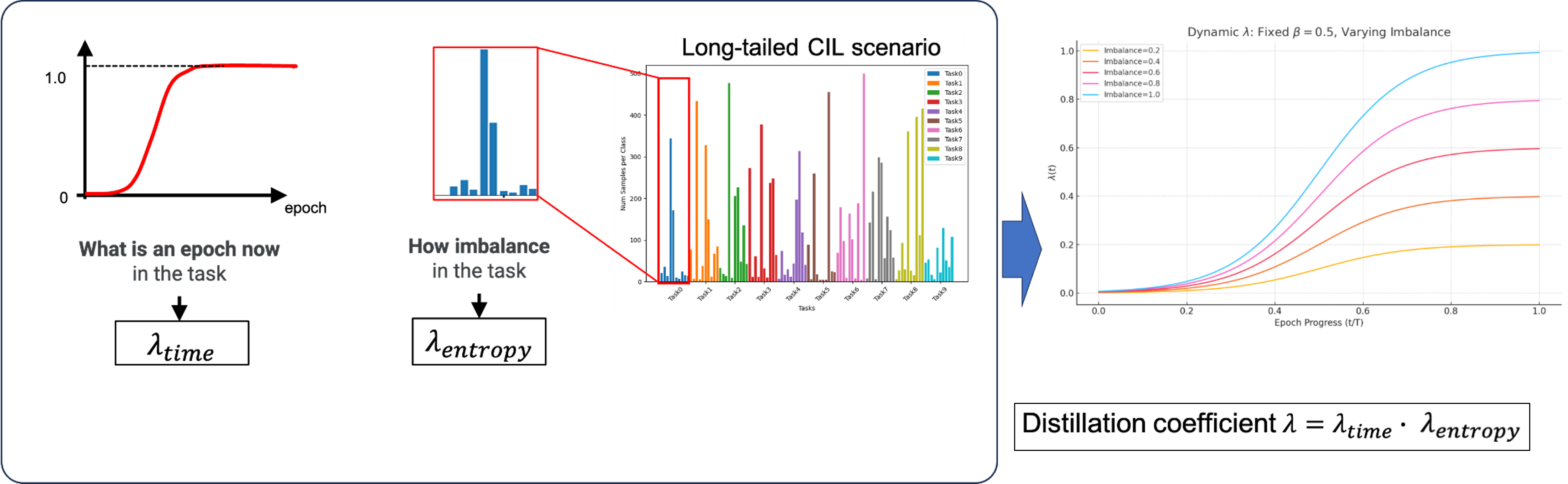}
    \caption{Conceptual Diagram for Deriving the Entropy-Aware Dynamic Distillation Coefficient $\lambda$. The left section illustrates $\lambda_{\mathrm{time}}$(time-based scheduling) which adjusts the distillation coefficient based on epoch progression within a task. The center section quantifies the imbalance in class distribution within a task and adjusts the distillation coefficient via $\lambda_{\mathrm{entropy}}$(entropy-based correction). The graph on the right visualizes the dynamic evolution of the final distillation coefficient $\lambda$ under various class imbalance levels (normalized entropy). The final distillation coefficient $\lambda$ is determined by multiplying $\lambda_{\mathrm{time}}$ and $\lambda_{\mathrm{entropy}}$.}
    \label{fig:method1}
\end{figure*}

% \begin{figure}[t]
%     \centering
%     \includegraphics[width=\linewidth]{lambda_schedule_entropy.png}
%     \caption{Visualization of temporal dynamics and class imbalance effects on the distillation coefficient $\lambda$. The evolution of $\lambda(t)$ is illustrated for various imbalance levels (normalized entropy), with $\beta$ fixed at 0.5.}
%         \label{fig:lambda_dynamic}
% \end{figure}

\subsection{Entropy-Aware Dynamic Distillation Coefficient}

Knowledge distillation is an effective technique in continual learning for retaining knowledge from previous tasks. However, excessive reliance on distillation can hinder the learning of newly introduced classes. Particularly under class imbalance, distillation tends to disproportionately preserve majority class representations, limiting the effective adaptation to minority classes.

To tackle this issue, the distillation loss weighting coefficient $\lambda$ is dynamically controlled by integrating training progression and the global class distribution imbalance.
{The time-based scheduling term $\lambda_{\mathrm{time}}$ and the entropy-based correction term $\lambda_{\mathrm{entropy}}$ are explicitly defined as follows.}
\begin{equation}
{\lambda_{\mathrm{time}} = \sigma\left(\frac{t}{T}\right)}
\end{equation}
\begin{equation}
{\lambda_{\mathrm{entropy}} = \mathcal{H}_{\mathrm{norm}} = -\sum_{k=1}^{K_{\mathrm{total}}} \frac{p_k \log p_k}{\log K_{\mathrm{total}}}}
\end{equation}
{The dynamic distillation coefficient $\lambda$ is subsequently determined by their multiplication.}
\begin{equation}
{\lambda = \lambda_{\mathrm{time}} \cdot \lambda_{\mathrm{entropy}}}
\end{equation}
where $t$ denotes the current epoch, $T$ is the total number of epochs within a task, and $\sigma(\cdot)$ represents the standard sigmoid function. This time-dependent term $\lambda_{\mathrm{time}}$ ensures that the influence of distillation gradually increases as training progresses, prioritizing new class learning in early stages and knowledge retention in later stages.

The term $\mathcal{H}_{\mathrm{norm}}$ represents the normalized entropy, which quantifies the degree of class imbalance. Unlike previous approaches that only consider the distribution of the current task, the normalized entropy is calculated based on the accumulated class distribution across all observed tasks to accurately capture the global imbalance. Here, $K_{\mathrm{total}}$ is the total number of classes observed so far, and $p_k$ represents the proportion of samples belonging to class $k$ in the accumulated data stream. $\mathcal{H}_{\mathrm{norm}}$ equals 1 when class distributions are uniform, and approaches 0 when highly imbalanced.

{Intra-task class imbalance is explicitly addressed by Gradient Reweighting (GR), rather than the entropy-based $\lambda_{\mathrm{entropy}}$. In this framework, GR assigns weights $w_c = \min_c\{G_c\} / G_c$ based on the cumulative gradient norm $G_c$. Smaller weights, where $w_c \ll 1$, are assigned to majority classes, whereas larger weights, where $w_c \approx 1$, are given to minority classes, directly equalizing intra-task parameter updates. The entropy-based $\lambda_{\mathrm{entropy}}$ is strictly reserved for managing inter-task distillation strength to prevent over-distillation of prior majority classes. These two mechanisms function complementarily.}

This formulation ensures that the maximum value of the distillation coefficient $\lambda$ is suppressed when the imbalance ratio is high (i.e., data is highly imbalanced). This reduction in distillation strength mitigates the bias towards majority classes and allows the model to concentrate more on learning features of new classes. Conversely, in balanced scenarios, $\lambda$ can reach higher values, encouraging the preservation of past knowledge.

~\cref{fig:method1} illustrates how $\lambda_{\mathrm{time}}$ and $\lambda_{\mathrm{entropy}}$ jointly determine the final distillation coefficient $\lambda$. The left side shows the time-based scheduling, and the center captures class imbalance via normalized entropy. The right visualizes $\lambda$ evolution under different imbalance conditions.

\subsection{Gradient Consistency Regularization for Stable Learning}

In continual learning, the class distribution of newly introduced tasks often differs from previous ones, leading to abrupt changes in the direction and magnitude of parameter updates. These rapid fluctuations can destabilize training, negatively impacting minority class learning. To mitigate this issue, conventional GR~\cite{He_2024_CVPR} aimed to dynamically adjust the weights of gradients for each class or task to promote minority class learning and alleviate overfitting to majority classes. However, GR could not account for the negative impact on minority classes caused by abrupt changes in the direction and magnitude of parameter updates across tasks.

Therefore, GCR is proposed, which enforces coherence between current gradients and their historical averages. Specifically, substantial deviations from the moving average of gradients computed in previous epochs are penalized.
The regularized gradient $\mathbf{g}_t'$ is defined as
\begin{equation}
\mathbf{g}_t' = \mathbf{g}_t + \lambda_{\mathrm{GCR}} (\mathbf{g}_t - \bar{\mathbf{g}}_{t-1})
\end{equation}
where $\mathbf{g}_t$ represents the current gradient, $\bar{\mathbf{g}}_{t-1}$ denotes the exponential moving average from previous epochs, and $\lambda_{\mathrm{GCR}}$ is a hyperparameter controlling the strength of regularization. 
The moving average is updated as 
\begin{equation}
\bar{\mathbf{g}}_t = \beta \bar{\mathbf{g}}_{t-1} + (1 - \beta) \mathbf{g}_t.
\end{equation}
The parameters were experimentally set to $\lambda_{\mathrm{GCR}} = 0.1$ and $\beta = 0.9$.
This regularization effectively reduces abrupt parameter shifts, ensuring consistent yet adaptable parameter updates, particularly beneficial for minority classes prone to unstable gradients. Thus, the proposed methodology simultaneously mitigates catastrophic forgetting and stabilizes incremental learning under class imbalance.

% TODO:

\section{EXPERIMENT}
\label{sec:Experiment}

\subsection{Experimental Settings}

\begin{figure}[t]
\centering
\includegraphics[width=0.75\linewidth]{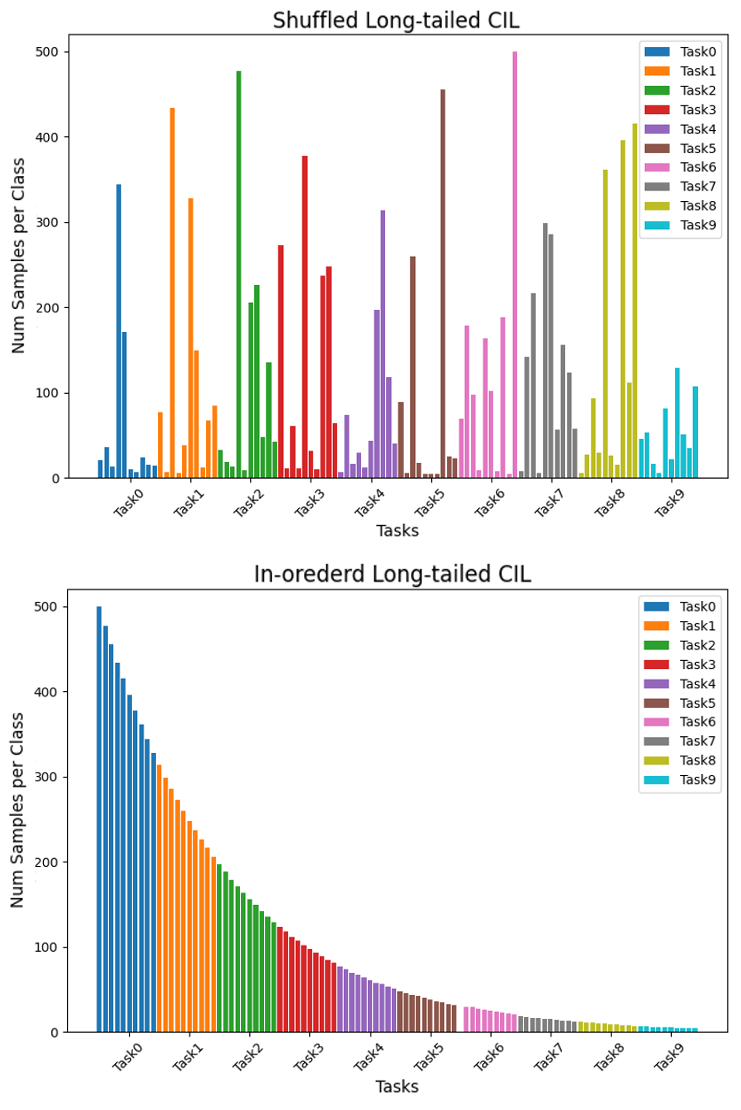}
\caption{Comparison of class orderings in Long-tailed Class-Incremental Learning. Top row illustrates the \textit{Shuffled} setting, with random mixing of classes across tasks, causing varying task difficulties. The bottom row shows the \textit{Ordered} setting, with tasks ordered from majority to minority classes.}
\label{fig:protocol}
\end{figure}

The proposed methodology was evaluated on three datasets: CIFAR-100-LT~\cite{Krizhevsky2009LearningML}, ImageNetSubset-LT~\cite{russakovsky2015imagenetlargescalevisual}, and Food101-LT~\cite{10.1007/978-3-319-10599-4_29}. For CIFAR-100-LT, an imbalanced dataset was constructed with an imbalance ratio of $\rho=100$, defined as the ratio between the largest and smallest classes. Each dataset was evenly divided into $N$ tasks, where $N \in \{10, 20\}$, and trained incrementally from scratch. A balanced distribution for test data was maintained to accurately measure performance under class imbalance.

ResNet-32 was employed for CIFAR-100-LT, and ResNet-18 was utilized for ImageNetSubset-LT and Food101-LT. The experimental conditions strictly follow those of GR.
For CIFAR-100-LT, training was conducted for 160 epochs, with an initial learning rate of 0.1, decayed by a factor of 0.1 at epochs 80 and 120. For ImageNetSubset-LT and Food101-LT, models were trained for 90 epochs, with an initial learning rate of 0.1, decayed by a factor of 0.1 at epochs 30 and 60. 
Optimization was performed using Stochastic Gradient Descent (SGD) with a batch size of 128.
To evaluate performance under long-tailed distributions, two class presentation orders were employed~\cref{fig:protocol}. The first setting, the \textit{Shuffled} setting, randomly distributes classes into tasks, leading to varied task difficulties due to mixing majority and minority classes. The second setting, the \textit{In-ordered} setting, arranges classes in descending order of sample count, progressively increasing difficulty as tasks advance.
Moreover, two training initialization scenarios were introduced. The \textit{From Scratch} scenario begins training from randomly initialized weights, learning all classes sequentially without prior knowledge. The \textit{From Half} scenario first learns half of the classes in an initial task, followed by incremental learning of the remaining half, assessing how effectively a partially pre-trained model acquires new knowledge.
Besides GR, iCaRL~\cite{rebuffi2017icarl} and PODNet~\cite{Douillard2020PODNetPO} were selected for comparison. Although these methods are not specifically designed for LT-CIL, they utilize distillation similarly to GR, making them appropriate baselines.

\subsection{Results}

\begin{table}[t]
    \centering
    \renewcommand{\arraystretch}{1.2}
    \caption{Average classification accuracy (\%) (From Scratch) after the final task.}
    \label{tab:methods_comparison_scratch}
    \resizebox{\columnwidth}{!}{
        \begin{tabular}{l|l|cc|cc}
            \toprule
            \textbf{Dataset} & \textbf{Method} & \multicolumn{2}{c|}{\textbf{In-ordered}} & \multicolumn{2}{c}{\textbf{Shuffled}} \\
            & & \textbf{10} & \textbf{20} & \textbf{10} & \textbf{20} \\
            \midrule
            CIFAR-100-LT & iCaRL & 20.5 & 18.6 & 21.4 & 19.8 \\
            & iCaRL+Ours & \textbf{22.6} & \textbf{20.2} & \textbf{21.8} & \textbf{19.9} \\
            & PODNet & 29.5 & 23.1 & 23.9 & 23.0 \\
            & PODNet+Ours & \textbf{29.6} & \textbf{25.0} & \textbf{24.3} & \textbf{23.1} \\
            & GR & 30.0 & 22.6 & 29.7 & 34.1 \\
            & GR+Ours & \textbf{35.0} & \textbf{26.8} & \textbf{33.5} & \textbf{36.2} \\
            \midrule
            ImageNetSubset-LT & iCaRL & 33.7 & 27.9 & 27.2 & 21.2 \\
            & iCaRL+Ours & \textbf{35.6} & \textbf{29.2} & \textbf{28.5} & \textbf{22.3} \\
            & PODNet & 33.9 & 28.7 & 27.6 & 22.3 \\
            & PODNet+Ours & \textbf{36.9} & \textbf{30.7} & \textbf{30.1} & \textbf{25.5} \\
            & GR & 38.7 & 39.3 & 34.2 & 35.3 \\
            & GR+Ours & \textbf{42.8} & \textbf{41.7} & \textbf{36.6} & \textbf{36.1} \\
            \midrule
            Food101-LT & PODNet & 19.0 & 15.2 & 20.6 & \textbf{19.7} \\
            & PODNet+Ours & \textbf{19.3} & \textbf{17.5} & \textbf{22.1} & 19.3 \\
            & GR & 33.5 & 30.1 & 21.5 & 24.0 \\
            & GR+Ours & \textbf{34.7} & \textbf{34.5} & \textbf{24.6} & \textbf{25.4} \\
            \bottomrule
        \end{tabular}
    }
\end{table}

\begin{table}[h]
    \centering
    \renewcommand{\arraystretch}{1.2}
    \caption{Average classification accuracy (\%) (From Half) after the final task.}
    \label{tab:methods_comparison_half}
    \resizebox{\columnwidth}{!}{
        \begin{tabular}{l|l|cc|cc}
            \toprule
            \textbf{Dataset} & \textbf{Method} & \multicolumn{2}{c|}{\textbf{In-ordered}} & \multicolumn{2}{c}{\textbf{Shuffled}} \\
            & & \textbf{5} & \textbf{10} & \textbf{5} & \textbf{10} \\
            \midrule
            CIFAR-100-LT & iCaRL & 21.2 & 20.4 & 21.6 & 21.0 \\
            & iCaRL+Ours & \textbf{23.4} & \textbf{21.2} & \textbf{21.9} & \textbf{21.6}  \\
            & PODNet & 30.0 & \textbf{29.6} & 35.6 & 35.5 \\
            & PODNet+Ours & \textbf{30.3} & 29.2 & \textbf{35.8} & \textbf{36.0} \\
            & GR & 31.4 & 26.5 & 37.4 & 34.3 \\
            & GR+Ours & \textbf{33.8} & \textbf{29.1} & \textbf{38.5} & \textbf{36.2} \\
            \midrule
            ImageNetSubset-LT & iCaRL & 38.0 & 36.7 & 36.1 & 35.5 \\
            & iCaRL+Ours & \textbf{38.8} & \textbf{38.2} & 36.3 & 35.8 \\
            & PODNet & 40.3 & 38.1 & 40.5 & 37.9 \\
            & PODNet+Ours & \textbf{41.6} & \textbf{39.2} & \textbf{40.7} & \textbf{38.1} \\
            & GR & 41.2 & 36.4 & \textbf{41.6} & 39.0\\
            & GR+Ours & \textbf{42.8} & \textbf{40.0} & 41.3 & \textbf{39.7} \\
            \midrule
            Food101-LT & PODNet & 34.9 & 29.3 & 33.1 & 29.0 \\
            & PODNet+Ours & \textbf{36.2} & \textbf{31.0} & \textbf{33.5} & \textbf{30.3} \\
            & GR & 37.7 & 31.1 & 36.4 & 30.5 \\
            & GR+Ours & \textbf{38.9} & \textbf{32.5} & \textbf{37.1} & \textbf{30.9} \\
            \bottomrule
        \end{tabular}
    }
\end{table}

~\cref{tab:methods_comparison_scratch} and ~\cref{tab:methods_comparison_half} show the final average accuracy after the last task under \textit{From Scratch} and \textit{From Half} scenarios, respectively.
~\cref{tab:methods_comparison_scratch} presents results for CIFAR-100-LT in the \textit{From Scratch} scenario. The "+Ours" designation indicates the integration of the proposed methodology with baseline approaches. The proposed framework consistently improved accuracy by up to approximately 3.6\%, reducing the negative impact of class imbalance while effectively preserving prior knowledge. In particular, the combination with GR (GR+Ours) demonstrated the most notable improvement, verifying the effectiveness of adaptive distillation.

~\cref{tab:methods_comparison_scratch} also details results for ImageNetSubset-LT and Food101-LT, confirming the efficacy of the methodology on larger-scale datasets and validating its general applicability.
Overall, results from ~\cref{tab:methods_comparison_scratch} show that the \textit{In-ordered} setting generally achieves higher accuracy than the \textit{Shuffled} setting. This trend is attributed to greater catastrophic forgetting in the \textit{Shuffled} setting, where data distributions vary unpredictably between tasks. Notably, for ImageNetSubset-LT, GR+Ours achieved the highest accuracy of 42.8\% under the \textit{In-ordered} setting. The results with PODNet+Ours indicate that the proposed framework can be effectively combined with other distillation-based approaches.

~\cref{tab:methods_comparison_half} presents results for the \textit{From Half} scenario. Although accuracy generally improved due to pre-training on initial classes, the proposed methodology consistently achieved performance improvements by efficiently leveraging prior knowledge while adapting to new classes.
Across both initialization conditions, the proposed framework consistently demonstrated improved performance. Particularly under \textit{From Scratch} conditions, despite starting without prior knowledge, the adaptive distillation strategy mitigated early overfitting, while GCR stabilized feature updating. Consequently, the framework proved effective even in practical scenarios lacking pre-trained models.

In the \textit{In-ordered} setting, existing methods typically develop biased representations toward majority classes presented first, limiting the recognition of subsequently introduced minority classes. However, the proposed approach maintains higher accuracy under this challenging scenario. This effectiveness is attributed to entropy-aware distillation, which attenuates initial bias, combined with GCR to preserve past knowledge.
Increasing the number of tasks reduces the classes per task, intensifying imbalance effects. Even under higher difficulty settings, such as $N=20$, the proposed framework maintained stable accuracy, demonstrating robust adaptability through dynamic distillation control and sustained knowledge retention via gradient consistency.

% ~\cref{fig:cifargraph}-~\cref{fig:imagenetgrraph} visualizes class-wise accuracy on CIFAR-100-LT and ImageNetSubset-LT. The horizontal axis shows the class index, and the vertical axis shows the accuracy (\%). Blue solid lines represent the baseline GR with 10 tasks (N=10), while red solid lines show GR with our proposed method for 10 tasks. Similarly, blue dashed lines are for baseline GR with 20 tasks (N=20), and red dashed lines are for GR with our method for 20 tasks. The curves for GR+Ours consistently outperform baseline GR across tasks, especially improving minority class performance. Although increasing the number of tasks (from 10 to 20) typically leads to performance degradation due to enhanced forgetting, our method significantly mitigates this effect.

\begin{figure}[t]
\centering
\begin{minipage}[t]{0.9\linewidth}
    \centering
        \includegraphics[width=\linewidth]{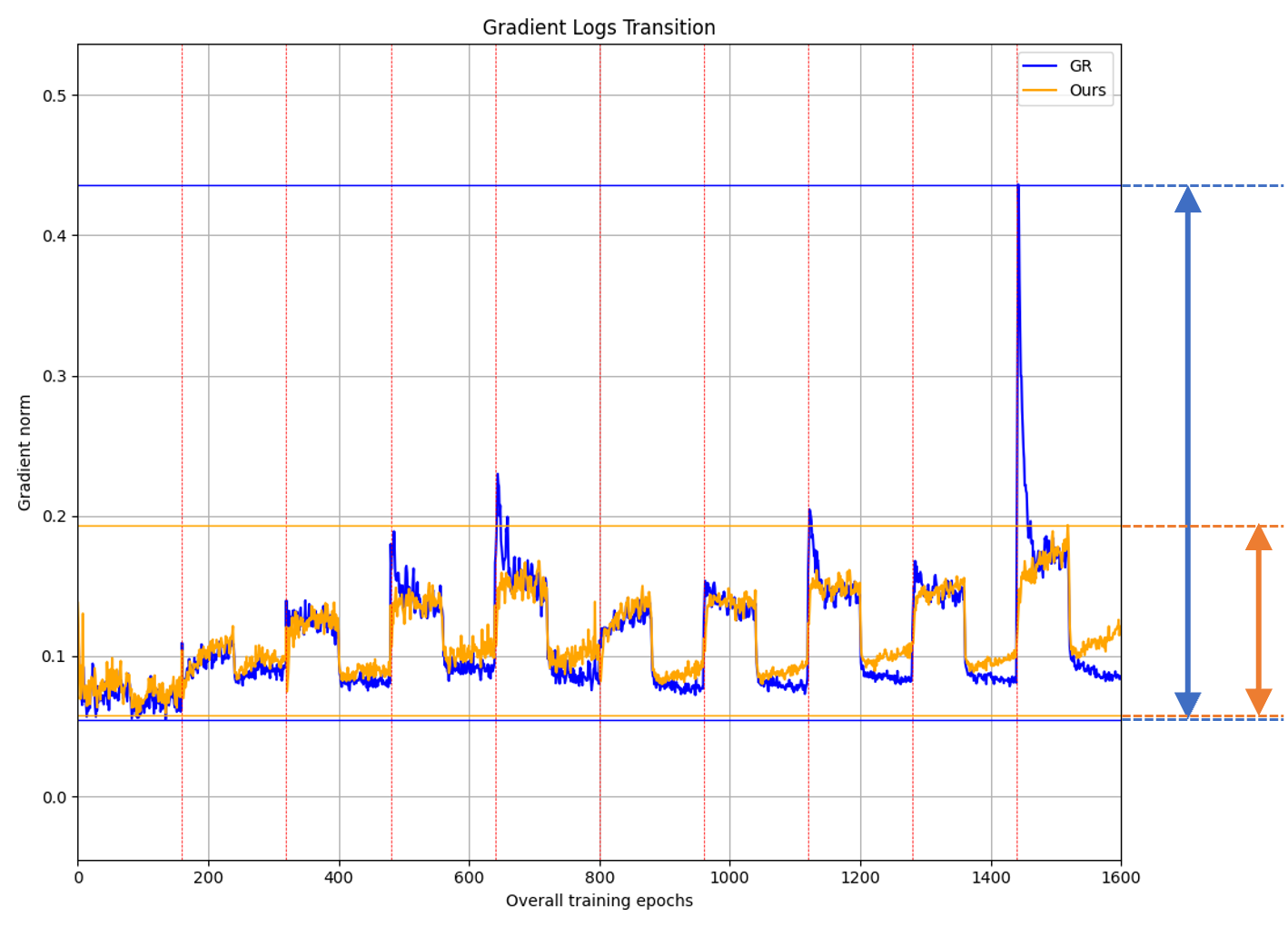}
        \subcaption{From Scratch}
        \label{fig:gradnorm}
\end{minipage}
\vspace{0.1cm}
\begin{minipage}[t]{0.9\linewidth}
    \centering
        \includegraphics[width=\linewidth]{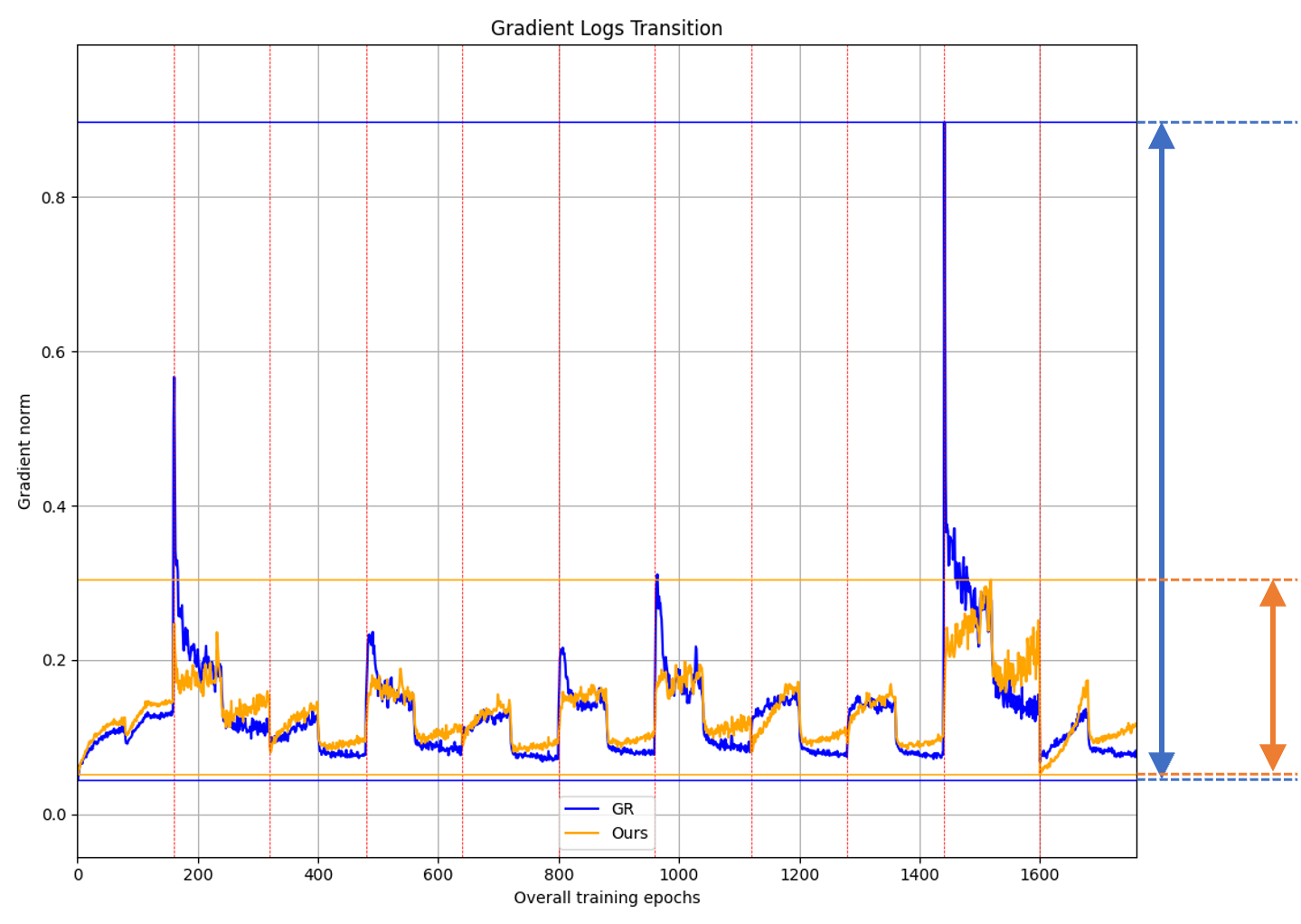}
        \subcaption{From Half}
        \label{fig:gradnorm2}
\end{minipage}
\caption{Gradient norm fluctuations during CIFAR-100-LT training with 10 shuffled tasks. The x-axis represents epochs, and the y-axis shows average gradient norms per mini-batch. Our method (orange curve) significantly reduces abrupt gradient fluctuations at task boundaries compared to the GR baseline (blue curve), indicating enhanced training stability. Horizontal lines show the min-max range of gradient norms within each epoch for each method.}
    \label{fig:gradnorm_combined}
\end{figure}

~\cref{fig:gradnorm_combined} illustrates the gradient norm fluctuations across epochs for CIFAR-100-LT. The horizontal axis represents the number of epochs, and the vertical axis shows the average gradient norm for each mini-batch within that epoch. Red lines mark the boundaries between tasks. The orange line represents the proposed methodology, and the blue line denotes GR. The horizontal lines for each method indicate the range between the maximum and minimum gradient norms within that epoch.
From this Figure, it is observed that the average gradient norm of GR sharply increases each time tasks switch, and the difference between maximum and minimum values becomes larger. This suggests that parameter updates are unstable due to knowledge interference between tasks.
In contrast, utilizing the proposed methodology, the average gradient norm per epoch remained relatively consistent, and no large fluctuations were observed during task transitions. The range between maximum and minimum values was kept narrow, confirming reduced learning variability.
This result indicates that GCR performs stable updates while maintaining consistency with past learning, yielding improvements in both accuracy and training stability.

\subsection{Ablation Study}

% \begin{figure}[t]
% \centering
% \includegraphics[width=0.8\linewidth]{cifar100.png}
% \caption{Class-wise accuracy on CIFAR-100-LT (In-ordered, From Scratch). The horizontal axis represents class indices, and the vertical axis indicates accuracy (\%). Solid lines correspond to experiments with 10 tasks ($N=10$), while dashed lines represent experiments with 20 tasks ($N=20$). Our method (GR+Ours, red) consistently outperforms the baseline GR (blue) for both task counts, notably reducing performance degradation for minority classes.}
% \label{fig:cifargraph}
% \end{figure}

% \begin{figure}[t]
% \centering
% \includegraphics[width=0.87\linewidth]{imagenetsubset.png}
% \caption{Class-wise accuracy on ImageNetSubset-LT (In-ordered, From Scratch). The horizontal axis represents class indices, and the vertical axis indicates accuracy (\%). Solid lines correspond to experiments with 10 tasks ($N=10$), while dashed lines represent experiments with 20 tasks ($N=20$). Our method (GR+Ours, red) consistently outperforms the baseline GR (blue). In all scenarios, the red lines (GR+Ours) consistently outperform the blue lines (GR), demonstrating that the proposed method achieves consistently higher classification accuracy than the baseline GR. This indicates the effectiveness of our approach.}
% \label{fig:imagenetgrraph}
% \end{figure}

\begin{table}[h]
    \centering
    \caption{Accuracy comparison based on distillation coefficient scheduling (CIFAR-100-LT From Scratch).}
    \label{tab:schedule}
    \resizebox{\columnwidth}{!}{
        \begin{tabular}{c|c|c|c|c}
            \toprule
            {\textbf{Protocol}} & \multicolumn{2}{c|}{\textbf{In-ordered}} & \multicolumn{2}{c}{\textbf{Shuffled}} \\
            \midrule
            \textbf{Tasks N} & \textbf{10} & \textbf{20} & \textbf{10} & \textbf{20} \\
            \midrule
            GR & 30.0 & 22.6 & 29.7 & 34.1 \\
            GR+Ours (Linear) & 30.4 & 24.9 & 32.5 & 31.4 \\
            GR+Ours (Sigmoid) & 30.1 & 24.0 & 31.7 & 30.9 \\
            GR+Ours (Entropy  $\times$ Linear) & \underline{33.5} & \underline{25.0} & \underline{33.0} & \underline{34.7} \\
            GR+Ours (Entropy  $\times$ Sigmoid) & \textbf{33.6} & \textbf{25.2} & \textbf{33.1} & \textbf{35.6} \\
            \bottomrule
        \end{tabular}
    }
\end{table}

~\cref{tab:schedule} compares accuracy under different distillation scheduling strategies on CIFAR-100-LT (From Scratch). 
In~\cref{tab:schedule}, "GR" is the baseline using a fixed coefficient. "GR+Ours (Linear)" and "GR+Ours (Sigmoid)" apply time-based scheduling, where the distillation weight increases linearly or with a sigmoid curve, respectively.
"GR+Ours (Entropy $\times$ Linear)" and "GR+Ours (Entropy $\times$ Sigmoid)" represent the proposed methodologies that combine time-based scheduling with entropy-based scaling to adjust for class imbalance.
This comparison highlights the individual and combined effects of temporal and distribution-aware control.

The results show that the proposed Entropy $\times$
Sigmoid scheduling combines temporal progression and class imbalance, significantly enhancing accuracy, especially in the challenging Shuffled setting (from 29.7\% to 33.1\%), confirming the effectiveness of our adaptive strategy.

{A class-wise accuracy analysis was conducted to demonstrate that Gradient Consistency Regularization (GCR) selectively stabilizes less frequent classes. The 100 classes of CIFAR-100-LT were categorized into Major, defined as exceeding 100 samples, Medium, and Minor groups. Compared to the GR baseline, the proposed methodology, comprising GR and GCR, improved accuracy in the Minor group by approximately 3\%, while preserving accuracy in the Major group. Since minority classes appear infrequently, the moving average $\bar{\mathbf{g}}$ functions as a restoring force in their absence, preventing excessive parameter drift toward majority classes.}

{Regarding computational efficiency, the proposed methodology results in a 1.3\% increase in training time and zero additional computational or memory overhead during inference. Because gradient consistency regularization and dynamic distillation only modify gradient updates and loss computation during training, inference overhead remains eliminated. These characteristics underscore the practical viability of achieving robust long-tailed incremental learning without extensive computational demands.}

% \begin{figure}[t]
%     \centering
%     \includegraphics[width=0.8\linewidth]{imagenet_forget.png}
%     \caption{Average classification accuracy (\%) along all tasks and overall forgetting rate (\%) on ImageNetSubset-LT under In-ordered setting. Comparison between GR and GR+Ours with different task splits (N=10, 20).
% }
%     \label{fig:imagenet_forget}
% \end{figure}

% \cref{fig:imagenet_forget} shows the Average classification accuracy along all tasks and average forgetting rate for the ImageNetSubset-LT dataset under the in-ordered task setting.
% The proposed method shows a lower forgetting rate in later tasks, which is opposite to the trend seen in existing methods. This result suggests that gradient regularization and adaptive distillation weights help the model learn new tasks without forgetting previous ones.
% We also measured the training time for "In-ordered", "From Scratch", and "10-task setting". The total training time for GR, GR+Ours (Sigmoid), and GR+Ours (Entropy $\times$ Sigmoid) was approximately 1850 seconds, 1870 seconds, and 1920 seconds, respectively, showing less than 3.8\% increase overall. This confirms that the proposed enhancements achieve better performance with only marginal additional computational cost.

\section{CONCLUSION}
\label{sec:Conc}

This study introduced an LT-CIL framework combining Gradient Consistency Regularization (GCR) and entropy-aware dynamic distillation. By stabilizing gradient fluctuations and adaptively balancing knowledge retention, the proposed methodology achieved up to a 5.0\% accuracy improvement, particularly on minority classes. Future work will focus on optimizing hyperparameters and refining distillation control strategies.

\bibliographystyle{IEEEbib}
\bibliography{refs}

\end{document}